\documentclass[10pt,twocolumn,letterpaper]{article}
\usepackage{wacv}
\usepackage{times}
\usepackage{epsfig}
\usepackage{graphicx}
\usepackage{amsmath}
\usepackage{amssymb}
\usepackage{booktabs}
\usepackage{array}
\usepackage{algpseudocode}
\usepackage{algorithm}
\usepackage{xcolor}
\usepackage{subfloat}
\usepackage{adjustbox}
\usepackage{multirow}
\usepackage{placeins}

\usepackage[pagebackref,breaklinks,colorlinks]{hyperref}

\usepackage[capitalize]{cleveref}
\crefname{section}{Sec.}{Secs.}
\Crefname{section}{Section}{Sections}
\Crefname{table}{Table}{Tables}
\crefname{table}{Tab.}{Tabs.}

\def\wacvPaperID{1467} 
\def\confName{WACV}
\def\confYear{2025}

\begin{document}



\title{Differentially Private Integrated Decision Gradients (IDG-DP) for Radar-based Human Activity Recognition}

\author{Idris Zakariyya
\and
Linda Tran
\and
Kaushik Bhargav Sivangi
\and
Paul Henderson
\and
Fani Deligianni\\
School of Computing Science, University of Glasgow, G12 8QQ, UK\\
{\tt\small\{idris.zakariyya, fani.deligianni\}@glasgow.ac.uk}
}



\maketitle
\thispagestyle{empty}


\begin{abstract}
    Human motion analysis offers significant potential for healthcare monitoring and early detection of diseases. The advent of radar-based sensing systems has captured the spotlight for they are able to operate without physical contact and they can integrate with pre-existing Wi-Fi networks. They are also seen as less privacy-invasive compared to camera-based systems. However, recent research has shown high accuracy in recognizing subjects or gender from radar gait patterns, raising privacy concerns. This study addresses these issues by investigating privacy vulnerabilities in radar-based Human Activity Recognition (HAR) systems and proposing a novel method for privacy preservation using Differential Privacy (DP) driven by attributions derived with Integrated Decision Gradient (IDG) algorithm. We investigate Black-box Membership Inference Attack (MIA) Models in HAR settings across various levels of attacker-accessible information. We extensively evaluated the effectiveness of the proposed IDG-DP method by designing a CNN-based HAR model and rigorously assessing its resilience against MIAs. Experimental results demonstrate the potential of IDG-DP in mitigating privacy attacks while maintaining utility across all settings, particularly excelling against label-only and shadow model black-box MIA attacks. This work represents a crucial step towards balancing the need for effective radar-based HAR with robust privacy protection in healthcare environments.
\end{abstract}

\vspace{-10mm}
\section{Introduction}
\label{sec:intro}


Demographic projections indicate that the worldwide population of individuals aged 60 and above will surge to approximately 1.4 billion people by 2030 ~\cite{Ageing_and_health}, leading to an increased reliance on advanced healthcare monitoring technologies in patients' homes \cite{GuReview2023}. Human motion analysis shows significant potential for healthcare monitoring and early disease detection ~\cite{Gait_Analysis_PD, Gait_Analysis_Mood}. Deep Neural Networks (DNNs) have exhibited remarkable efficacy in Human Activity Recognition (HAR) for monitoring patients and detecting abnormalities \cite{gul2020patient,beddiar2020vision}. Especially, in home settings these technologies can play a key role in preventive and proactive healthcare strategies by enabling personalised care systems ~\cite{wassan2024deep,GuReview2023}.  


However, while DNNs excel at encoding input features, this capability also renders them vulnerable to privacy breaches ~\cite{liu2020privacy,Jegorova2023}.
For instance, Membership Inference Attacks (MIA) \cite{rigaki2023survey, humphries2023investigating, shokri2017membership} and Model Inversion (MI) attacks ~\cite{he2019model} can disclose private information about the patients and their training data by accessing only the pre-trained model outputs. 
To mitigate such risks, privacy-preserving techniques needs to be incorporated into HAR systems to safeguard sensitive user information. Differential Privacy (DP) is a commonly used technique to enhance privacy in deep learning models ~\cite{chilukoti2022privacypreserving, fujimoto2023differential}. DP quantifies the risk of an individual's information being disclosed by ensuring that the model's output is not significantly affected by the inclusion or exclusion of any single individual's data. The implementation of DP involves introducing  controlled perturbations to the data.  

Recently, human motion sensing with radar emerges as a prominent sensing technology for continuous monitoring thanks to its non-intrusive nature ~\cite{Radar_Promising}. This makes it suitable for privacy-sensitive environments such as assisted-living facilities, hospitals and homes. Nevertheless, recent studies have revealed high accuracy in subject recognition from radar human gait patterns, challenging the perception of privacy in radar-based systems  ~\cite{rs12142237, 9455218}. This underscores the need for implementing privacy safeguards in human motion sensing systems, regardless of whether the output is visually identifiable by humans. 

Previous research in human motion analysis has primarily focused on privacy preservation of Red Green and Blue (RGB) videos on a frame by frame basis ~\cite{fitwi2021enforcing}. Some methods use anonymisation techniques by replacing the face or whole body with synthetic data\cite{hukkelaas2019deepprivacy}, while other approaches use obfuscation of sensitive attributes \cite{wang2021infoscrub}. In most cases, the utility of the data is severely compromised, hindering the application of these methods in healthcare. More recently, it has been recognised that other human motion tracking modalities such as accelerometer, gyroscope sensor data and radar data also record sensitive information that can help identify users and track them during their daily activities ~\cite{fujimoto2023differential,Isawa17,rs12142237, 9455218}. However, to our knowledge, there is no systematic work modeling threats and evaluating the robustness of privacy preservation techniques under sophisticated machine learning (ML)-driven attacks in radar-based systems. 


Here we firstly define black-box MIA that are relevant to HAR setting with radar data. We assume that the adversary has access to the logit space of the model and it is also possible to gain partial access to the training data or their underlying distribution. In these ways, the adversary can orchestrate attacks to identify individuals. 
Subsequently, we propose a novel method, named IDG-DP, based on DP and the Integrated Decision Gradient (IDG) attribution algorithm ~\cite{walker2024integrated}. The IDG was selected for its ability to compute attributions precisely at the model's decision points, making it a superior attribution algorithm with enhanced performance ~\cite{walker2024integrated}. IDG-DP injects more noise in input features that contribute more towards the subject identification than to the activity recognition and thus it preserves privacy, while it maintains high performance in HAR. 


Our paper presents the following contributions: 

\vspace{-2.5mm}
\begin{enumerate}
\itemsep-0.1em 
    \item To our knowledge, we are the first to systematically investigate state-of-the-art threat models that are relevant to HAR in a home setting with radar technology. This enables the identification of sensitive information leakage that is not perceptible to a human observer. 
    \item We introduce a novel methodology that drives DP by identifying the model's highest attributions during training. In this way, we achieve a better balance between data utility and privacy preservation.
    \item We devise a rigorous evaluation strategy of the mitigation capabilities of the proposed approach against black-box MIA attacks.
\end{enumerate}
\vspace{-3mm}
We exploit a publicly available dataset on HAR with radar data ~\cite{fioranelli2019radar} to assess the effectiveness of the proposed IDG-DP privacy method under various black-box MIA attacks. We demonstrate promising results in balancing data utility and privacy in the data.


\section{Background and Related Work}

\subsection{Human Activity Recognition with Radar Technology}
Human motion analysis involves examining human movement patterns, employing diverse technologies that involve multiple ~\cite{liu2016human} and single camera setups ~\cite{wang2024markerless}. Advancements in pose extraction, human tracking and human activity recognition (HAR) have made the technology accessible to patients at home and suitable for continuous monitoring ~\cite{6473854, s21093158, bdcc3010003}. Nevertheless, applying these techniques broadly in healthcare is met with significant hurdles, including ethical considerations, safety, and privacy issues, all of which must be thoroughly considered and resolved prior to widespread adoption ~\cite{heitzinger2021foundation}. Other technologies, such as wearables have been suggested for the analysis of body motion patterns but they require complex setup that exclude their use in patients with cognitive impairments ~\cite{nahavandi2022application}.  

Radar sensors can detect human motion signatures through the analysis of frequency modulations in the radio-frequency spectrum, which are induced by the reflection of signals from wifi/radar devices off moving targets \cite{yang2023human}. Furthermore, radar sensing is not affected by variable lighting conditions and eliminates the need for individuals to wear any sensors. Radar sensing has attracted attention because it can promote acceptance of monitoring technologies in healthcare applications, since it does not produce images that allow for the immediate recognition of individuals and their settings by human observers \cite{li2019survey}. 
Within radar systems, human motion is identified by analyzing micro-Doppler frequency shifts caused by vibrations, translational motions, and rotations of body parts ~\cite{ding2019continuous}. Various DNNs architectures including Convolutional Neural Network (CNN) ~\cite{8307105, 7314905, 8476568,9385952}, Recurrent Neural Network (RNN) ~\cite{8249173, 9130759} and auto-encoders ~\cite{8283539} have been employed for radar-based HAR ~\cite{li2019survey}.

\subsection{Privacy in Human Motion Analysis}

\subsubsection{Biometrics in Human Motion}\label{biometrics}
Radar technology enables innovative home monitoring applications, but raises privacy concerns due to continuous surveillance. Beyond data theft, real-time tracking of individuals at home poses greater risks than traditional computer vision applications used for diagnostics.
Whereas video data can reveal subject's identity directly, privacy preservation against threats that expose subject identity via her/his body movements are less studied. Individuals exhibit distinct movement patterns that can be used as biometrics ~\cite{makowski2020biometric}. These identity signatures are also present in radar technologies ~\cite{Cutting_Kozlowski_1977,Saleem24} and even wearables ~\cite{blasco2016survey,  zhao2021robust, pourbemany2023survey}. 
Leveraging this knowledge enables effective human recognition ~\cite{das2017design, mohsen2019authentication, behera2021robust} but it also creates concerns about users privacy under sophisticated ML driven attacks. The majority of subject recognition methods focus on analyzing the gait motion, with walking or running consistently yielding the most successful outcomes ~\cite{Yang2019}. 

Although, manually extracted features have demonstrated effectiveness ~\cite{kuehne2011hmdb}, deep CNN models have developed to automatically extract features from micro-Doppler signatures for subject recognition ~\cite{Yang2019, li2020hierarchical}. Transfer learning, employing pre-trained AlexNet on ImageNet dataset, has proven to be effective in subject recognition based on CNN models ~\cite{9455218}. For single subject recognition tasks, Generative Adversarial Networks (GANs) were also utilized, where the discriminator functions as a one-class classifier and the generator generates adversarial samples to aid in training ~\cite{Ji2021}. This approach could be particularly useful in identifying specific people and removing non-consented subjects from the scene ~\cite{soleimani2021cross}.

\subsubsection{Privacy Threats Against DNN Models in HAR }\label{privacy_threats}
DNNs occasionally assign high likelihood to some input samples, which reflects memorisation of the training data and it can lead to inadvertently disclose sensitive information without user consent~\cite{Isawa17}. An adversary can have access to the model's architecture and weights and/or full or partial access to the training data. Sophisticated attacks can be orchestrated even when the attacker does not have access to the model's weights and parameters \cite{mireshghallah2020privacy}. For example, in MIA an adversary with access to the model's logit space might attempt to infer whether a particular data record belongs to a subject whose data has been used for training \cite{rigaki2023survey, humphries2023investigating}. In HAR, this might have severe implications, resulting in identifying people and tracking them while they perform different activities at home.   
Despite their apparent simplicity, MIA form the basis for more robust extraction attacks ~\cite{rigaki2023survey}. The accurate identification of users within sensitive datasets constitutes a serious privacy breach. 
\subsubsection{Defense methods against MIA}
Defenses against MIA \cite{hu2023defenses} fall into four categories: Regularization, Transfer Learning, Generative Models-based, and Information Perturbation. Regularization techniques reduce overfitting and improve generalization, making models less vulnerable to MIA by adjusting internal parameters during training (e.g., L2 norm, dropout, and early stopping) or data augmentation and label smoothing \cite{kaya2020effectiveness}\cite{yeom2018privacy}\cite{wang2020against}. 
Generative models like VAEs, GANs, and EBMs produce alternative datasets that mimic the original data distribution, reducing membership information leakage by decoupling the original data from the model\cite{chen2021protect}\cite{hu2022defending} while maintaining overall data characteristics for effective training. Transfer learning protects against MIA\cite{papernot2016semi} by leveraging knowledge from different domains, reducing the need for target data access.  
While these methods offer some protection, they lack theoretical guarantees for the privacy of underlying data or models.  
Consequently, information perturbation techniques have gained popularity due to their stronger theoretical foundations. These approaches defend againt MIA by adding noise and include differential privacy\cite{humphries2023investigating}, output perturbation (adjusting confidence scores)\cite{jia2019memguard}, and data perturbation (hiding member information)\cite{hu2023defenses} to obscure sensitive data while balancing utility. 
 
\subsubsection{Privacy-Preservation techniques in HAR }
Privacy preservation technologies in human motion analysis have mainly tackled video cameras due to their inherent nature to directly expose the subject identities, ie. facial features and surrounding space ~\cite{yang2022secure,Hukkelas2023}. Preservation of privacy for these video cameras has been implemented via traditional techniques that involve blurring, pixelation and distortion. More recently more advanced `anonymisation' techniques have emerged with learnable optics and generative AI ~\cite{lopez2024privacy}. Learnable optics anonymises the data by removing privacy-breaching attributes while retaining the system's main capability ~\cite{9710270, dave2022spact, Hinojosa_2022}. The concept involves optimizing the lens to effectively remove sensitive attributes or degrade video quality. For instance, authors in ~\cite{9710270} optimizes an optical hardware encoder with a software decoder CNN to degrade private attributes of people filmed on videos while maintaining important features to perform human pose estimation. Although, these approaches offer some protection, since images cannot be identified by human observers, there is no evidence that they can withstand more sophisticated attacks that are eminent in healthcare applications. 

DP has been originally proposed by ~\cite{Dwork06} and it provides a strong theoretical privacy guarantee while analyzing datasets. Normally, it involves injecting calibrated noise (typically quantified by privacy budget $\epsilon$) into the input features, the gradients computation and/or the model parameters. Although, the risk of identifying or learning specific information about any individual sample is minimised, data utility might be severely compromised ~\cite{Yang_Lyu_Zhao_Zhu_Lam_2020}. 
Furthermore, most techniques focus on frame by frame analysis and there is less work on the privacy-preservation of the spatio-temporal dynamics present in whole video sequences \cite{li2023stprivacy}.


\section{Methods} \label{method}

The method section begins by detailing the underlying threat models that our method addresses (Section  \ref{threat_model_section}). This provides the context for understanding the privacy risks in radar-based HAR systems. We then describe the HAR model used in our study (Section \ref{multi-task_model}. This section explains the architecture and functionality of the model that forms the basis of our privacy-preserving approach. Finally, we introduce our novel IDG-DP technique (Section \ref{idg_dp}). This section elaborates on how we combine IDG with DP to mitigate the identified privacy risks while maintaining the utility of the HAR model.



\subsection{Problem definition and Threat Model} \label{threat_model_section}

We exploit black-box MIA attacks to understand and tackle system's vulnerabilities in DNNs used in computer vision for HAR, because they can be executed with different levels of adversarial knowledge, their effectiveness can be estimated by common evaluation metrics and they can be used as the basis for several other attacks that involve reconstruction of input data, injection of malicious information to degrade model's performance and model stealing attacks. Therefore, black-box MIA attack models will serve as the basis for assessing the mitigation capability of a privacy preservation technique in HAR~\cite{jayaraman2019evaluating}. Specifically, we describe the MIA according to Definition \hyperref[Def_1]{1} ~\cite{art}.
\vskip4pt \noindent

\textbf{Definition 1} \label{Def_1}: \textit{Consider an adversary $\mathcal{A}$, who is provided with a specific data point $z$, the distribution $\mathcal{D}$ from which a training set $\mathcal{S}$ has been drawn, and access to a model $\mathcal{M}$ trained on $\mathcal{S}$. The objective of $\mathcal{A}$ is to ascertain whether data point $z$ was included in the dataset $\mathcal{S}$ used to train model $\mathcal{M}$. This process of determining the membership of $z$ in $\mathcal{S}$ constitutes MIA. ~\cite{Carlini21, Yeom17}}. 

\vskip2pt \noindent

\begin{figure}[t]
\centering
\includegraphics[width=0.47\textwidth]{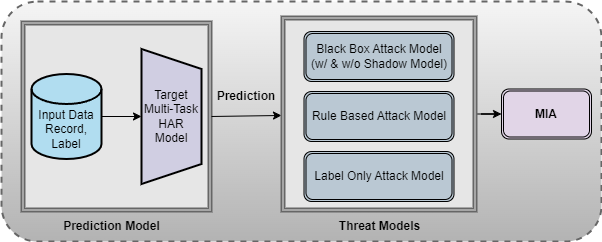}
\caption{Investigated Threat Models}
\label{fig:miatm}
\end{figure}

The investigated MIA threat models  are shown in Figure \ref{fig:miatm}. These attack models are defined below:

\vspace{-2.5mm}
\begin{enumerate}
\itemsep-0.1em 
    \item \textbf{Black-Box Attack Setting}: Under this setting, the adversary $\mathcal{A}$ only has access to the outputs of $\mathcal{M}$ when provided with inputs. $\mathcal{A}$ queries $\mathcal{M}$ with datapoint $z$ and receives prediction label. The adversary utilizes an attack model $\mathcal{A}_m$ and  assesses whether $z$ was used in training of $\mathcal{M}$.  

    \textbf{Shadow Model Black-Box Attack Setting}: In this scenario, the adversary $\mathcal{A}$ initializes one or more shadow models $\mathcal{M}_s$ based on target model $\mathcal{M}$. The adversary then generates attacks datasets derived from the statistical distribution of the original dataset $\mathcal{S}$.    

    \item \textbf{Rule Based Attack Setting}: Under this setting, we assume a simple decisionrule as a naive baseline attack method. The adversary $\mathcal{A}$ assumes that if $\mathcal{M}$ accurately predicts the label of data point $z$, then it is interpreted as evidence that $z$ is a member of $\mathcal{S}$ \cite{luqman2023membership}.

    
    \item \textbf{Label-Only Attack Setting}: Under this setting, adversary $\mathcal{A}$ works under the constraint of only having access to the class label outputs of $\mathcal{M}$ when queried with $z$. Analyzing the discrete label $l$ (without probabilistic/ confidence information), $\mathcal{A}$ assesses whether $z$ was part of $\mathcal{S}$.

\end{enumerate}

In other words, in the black-box attack ~\cite{shokri2017membership}, the attacker uses inputs from a specific target model to train an attack model, that can be use to check whether a specific data sample is  in the dataset. Unlike the black-box attack ~\cite{shokri2017membership}, the black-box rule-based does not require fitting a training model, rather it uses certain rules to check whether a data sample is in a dataset.  The black-box ~\cite{shokri2017membership} and rule-based attacks are evaluated using both a subsection of the training and a set of unseen to the model data records, with total accuracy serving as the metric to assess information leakage.  
Additionally, the label-only attack ~\cite{choquette2021label} is more general and requires only the hard label of a trained model without the prediction scores.
The attacker in the label-only attack ~\cite{choquette2021label}, assesses various perturbation measures to identify the noise level needed to change the classifier's prediction for a given sample, with the intuition being to perturb the input sufficiently to cause model misclassification. Experimental codes are  accessible at ~\footnote{\url{https://github.com/izakariyya/IDG-DP/tree/main}}. 

\subsection{Multi-Task Network} \label{multi-task_model}
As demonstrated in ~\cite{Ni2020HumanIB}, transfer-learned ResNet ~\cite{resnet} models are successful at gait-based person classification. We adapt this architecture to create a multi-task network which is capable of identifying both activities and personnel from the Micro-Doppler signatures. The Micro-Doppler signatures were created based on the Fast Fourier Transform (FFT) of overlapping time windows. Table  \ref{architecture} in Appendix \ref{multi-task-arch} illustrates the architecture of our network. This multi-task model functions as an evaluation baseline to the efficacy of the proposed privacy preservation technique. 

\subsection{Integrated Decision Gradients Differential Privacy (IDG-DP)}\label{idg_dp}
 We develop a novel privacy preservation approach, namely IDG-DP, that builds on both IDG and Pure-DP method proposed in ~\cite{dwork2014algorithmic}. 
 IDG leverages that moving across a gradient path from the decision regions would rapidly affect the output logit \cite{walker2024integrated}. Its robustness stems from its strong theoretical guarantees of sensitivity, implementation invariance and completeness.
 
 Pure-DP method provides a way to control the impact of an individual’s data on the outcomes of computations of ML models by intentionally adding noise into the data. Specifically, we use $\epsilon$-DP (Definition  \hyperref[Def_2]{2}) with a Laplace  mechanism ~\cite{fernandes2021laplace} as the randomised algorithm. The Laplace DP noise mechanism adds noise sampled from the Laplace distribution to the data. This method is known for providing a good balance between privacy protection and data utility, approaching optimal utility in many DP applications ~\cite{dwork2014algorithmic, fernandes2021laplace} than the Gaussian distribution which only satisfies the ($\epsilon, \delta$) DP with $\epsilon < 1$ ~\cite{dwork2014algorithmic}. The chosen Laplace distribution consists of mean 0 and and a scale parameter $b=\frac{\Delta f}{\epsilon}$. where $\Delta f$ is the sensitivity of the query indicating the maximum change in the query output resulting from the addition or removal of a single data point. This sensitivity informs the required noise level, while $\epsilon$ serves as a privacy parameter, with smaller value imply stronger privacy guarantees.

\vskip8pt
\noindent\textbf{Definition 2}\label{Def_2}: \textit{Let $\epsilon>0$, a randomized algorithm $M$ guarantees $\epsilon$-differential privacy if for all neighboring input datasets $D_1$ and $D_2$ differing on at most one element and $\forall S  \subseteq Range(M)$, we have ~\cite{dwork2014algorithmic}:}
\[Pr[M(D_1) \in S] \leq e^\epsilon Pr[M(D2) \in S]\]

The DP techniques often come at the cost of reducing data quality and utility ~\cite{Yang20}. Hence, drawing inspiration from ~\cite{fujimoto2023differential} we propose to add  stronger noise only to features that are important for subject recognition and less noise to other features. To identify which features contribute mostly to the output decision, we adapt the IDG approach ~\cite{walker2024integrated}. IDG builds on Integrated Gradients (IG), which fulfil strong axiomatic properties ~\cite{sundararajan2017axiomatic}. However, the IDG assigns larger weights to path integrals from regions where the gradients do not suffer form saturation effects and thus reflect stronger impact on the model's decision.


The procedure for obtaining IDG-DP is detailed in Algorithm \ref{alg:dp_model}. The Function DP in Algorithm \ref{alg:dp_model}, require a dataset and an instance of the multi-task HAR model with an $\epsilon$ and an attribution threshold that control the addition of noise. Initially, the attribution maps for both the subjects and activity recognition tasks are estimated using the IDG ~\cite{walker2024integrated} technique. Then, lines 4 and 5 of Algorithm \ref{alg:dp_model} estimated the average attributions of activities and subjects to discern those features and inject noise based on their importance. 


\begin{algorithm}
\small
\caption{Algorithmic description of \textbf{IDG-DP}}
\label{alg:dp_model}
\hspace*{\algorithmicindent} \textbf{Input:} \{$\epsilon$, attribution threshold $\textit{at}$, training data $ \mathcal D$, number of iterations $\mathcal T$, multi-task HAR ($\mathcal {MHAR}$)\}

 \hspace*{\algorithmicindent} \textbf{Output:} \{$\mathcal {\textbf{IDG-DP}}$\}

\begin{algorithmic}[1]

\State \textbf{Function} \textsc{DP}($\mathcal D$, $\mathcal {MHAR}$)
    \State estimates $sa$  \Comment{$sa$ = subjects attribution feature maps}
    \State estimates $aa$   \Comment{$aa$ = activity attribution feature maps}
    \State estimates avg of $aa$ and $sa$   \Comment{avg = average}
    \State estimates std of $aa$ and $sa$   \Comment{std = standard deviation}
    \State estimates $ns$ based on $at$ \Comment{ns = noise indices = indices where avg $<$ \textit{at}}
    \For{$i = 0$ to $\mathcal T$}
        \State $ \mathcal {NM} =  \mathcal {MHAR}$($\epsilon$, $ns$) \Comment{$NM$ = noise model}
        \If{$ \mathcal {NM}$ converges}
            \State $\mathcal {NM} = \mathcal {\textbf{IDG-DP}}$
            \State \textbf{break}
        \EndIf
    \EndFor
    \\
    \Return $\mathcal {\textbf{IDG-DP}}$
\State \textbf{End Function}

\end{algorithmic}
\end{algorithm}

\begin{figure}[t]
\begin{center}
\includegraphics[width=\linewidth]{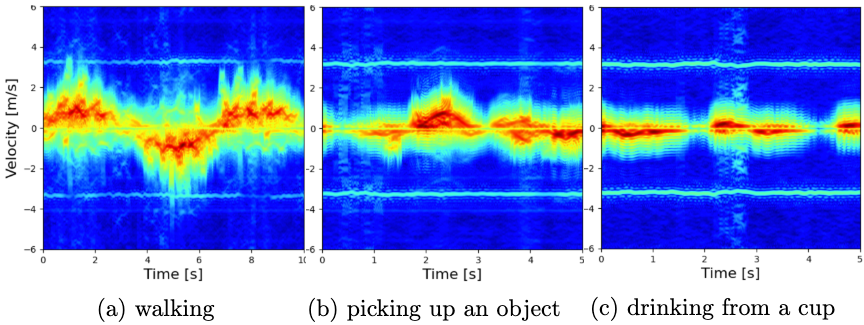}
\end{center}
\vspace{-7mm}
\caption{\textbf{Micro-Doppler spectrograms of a person performing 3 different activities.} Unique movement patterns visible in micro-Doppler spectrograms enable both activity and subject recognition.}
\label{fig:m_D activities}
\end{figure}

\vspace{-5mm}
\section{Evaluation}
\subsection{Dataset and preprocessing}
The dataset used in this study is the publicly available Radar dataset collected in 2019 at the Age UK West Cumbria centre ~\cite{fioranelli2019radar}. The radar used was a Frequency Modulated Continuous Wave (FMCW) radar operating at 5.8 GHz with 400 MHz bandwidth and 1ms chirp duration ~\cite{fioranelli2019radar}. The raw-data are temporal series organised into a matrix according to the Pulse Repetition Frequency (PRF). Data are converted to micro-doppler spectrograms normalised in log scale by estimating the FFT over overlapping windows. Figure \ref{fig:m_D activities} shows the micro-doppler signatures used for classification whilst considering three activities namely: walking, picking up an object and drinking from a cup.
We consider ten subjects that have repeated all the activities more than once. This provided with the total number of 90 sessions, where 60 samples are used for model training and the remaining 30 data samples are used for testing.

\subsection{Experimental Setup}

Experiments were run on a NVIDIA RTX A5000 GPU. All models were developed using Pytorch library ~\cite{paszke2019pytorch} whilst the Adversarial Robustness Toolbox (ART) ~\cite{nicolae2018adversarial} was utilized in generating the attacks defined in Section \ref{threat_model_section}.


\begin{figure*}
\small
\centering
\subfloat[\:{$\epsilon$ = 0.5, 1.0, 1.2, 1.5, 2.0, 2.5}]{\includegraphics[width=0.27\textwidth]{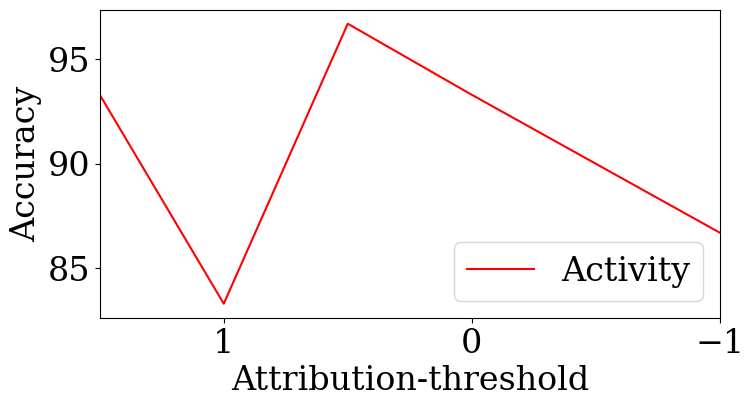}}
\subfloat[\:{{Attribution Threshold = $0.00025$}}]{\includegraphics[width=0.27\textwidth]{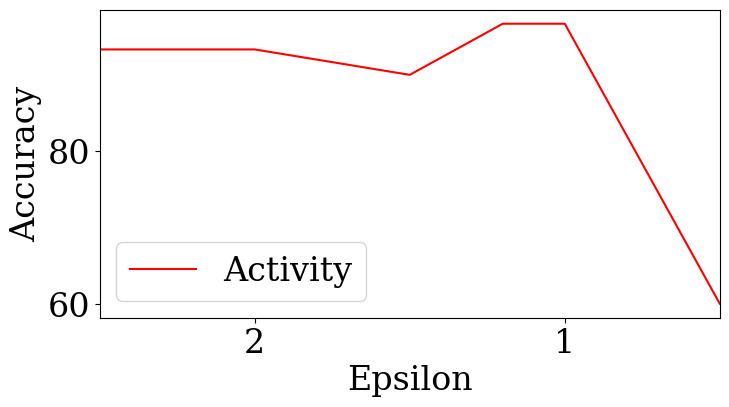}}
\subfloat[\:{{Pixel Attribution for Activity}}]{{\includegraphics[width=0.27\textwidth]{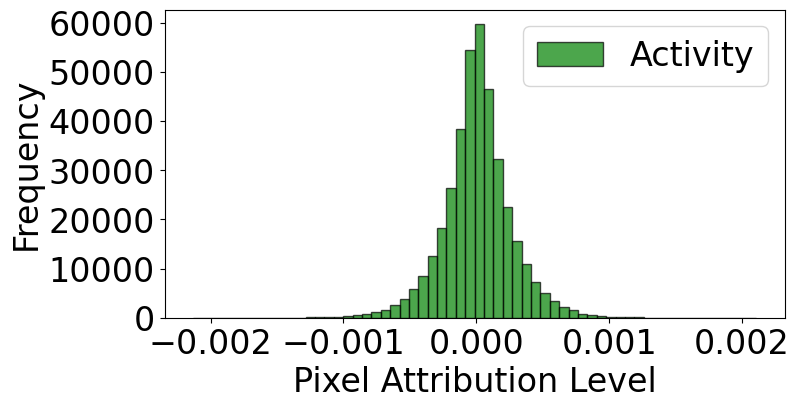}}}
\caption{\textbf{Impact of Different Activity Attribution Thresholds and  $\epsilon$ Levels on a 10-Subject HAR Model.} (a) {average accuracy of HAR across $\epsilon$ values while the attribution threshold varies}, (b) { Accuracy of HAR across  $\epsilon$ values demonstrates the utility-privacy trade-off of the proposed method}, (c) {Distribution of pixel attributions derived from IDG}. }
\label{fig:igd_dp_noise}
\end{figure*}

\subsection{Multi-task HAR DP Noise Models Implementation Procedure}

Section \ref{multi-task_model} provides details on the HAR model input, architecture and Secton \ref{implem_details} in Appendix described the training processes. 
To determine the effective $\epsilon$ and attribution threshold for the proposed IDG-DP method, an initial investigation was conducted. Figure~\ref{fig:igd_dp_noise} (a) shows the impact of various $\epsilon$ and attribution threshold values on the performance of user and activity recognition using the multi-task HAR model. Figure~\ref{fig:igd_dp_noise}(b) illustrates the recorded performance across various $\epsilon$ values ranging from 2.5 to 0.05 with 15 steps. Lower epsilon values (indicating stronger privacy), make it less likely for a user to be identified. Conversely, a higher HAR accuracy indicates better performance of the activity recognition model. At $\epsilon=0.5$, the subject accuracy with DP is 10\%, while the activity recognition accuracy is 40\%. 
The balance between HAR performance and robustness against attacks was determined by examining the impact of varying pixel attribution level thresholds and the DP parameter $\epsilon$ on the performance of the HAR model. Figure \ref{fig:igd_dp_noise} (c) shows the histogram of pixel attribution for users and activity. Based on these findings, the DP noise model was constructed to inject the appropriate Laplace noise to maintain the performance of HAR. Therefore, it was found that $\epsilon = 1.20$ and an attribution threshold of 0.00025 work best for achieving a good compromise between optimal performance and privacy. 

To compare the proposed approach, we constructed several other attribution-based DP models, namely IG-DP, Integrated InputX Gradient (IIG) ~\cite{sacha2017dynamic} IIG-DP, Integrated Smooth Gradients (ISG) ~\cite{walker2024integrated}  ISG-DP, Saliency (Sal) ~\cite{simonyan2013deep} Sal-DP, and Grad-Cam (GradC)~\cite{GradCam} GradC-DP. All these models use the same $\epsilon = 1.20$ and attribution threshold equal to 0.00025. The IG attributions were computed using a step size of 50 and an internal batch size of 100. For the computations of ISG attributions, a noise tunnel was introduced into the IG attribution to smooth the gradients, using 2 samples for injecting the smooth Gaussian noise with an internal batch size of 1. The IDG attributions were computed using a batch size of 25 and 50 steps. 

Additionally, we benchmark our approach against a {baseline model with DP (Base-DP) as described in Section ~\ref{base_dp_performance} in Appendix} and  DP-based learnable mask model, which resembles the principles behind optimized optics ~\cite{9710270, dave2022spact, Hinojosa_2022}. This `optics' model is constructed using a single convolutional layer, tuned by a learnable parameter per pixel to regulate the level of mask noise introduced.




\subsection{Attack procedures} \label{attacks_procedure}
 To create MIA black-box attack ~\cite{shokri2017membership}, we create an attack model using an ART ~\cite{nicolae2018adversarial}. The attack model was trained using a subset of the training and testing data from the radar dataset ~\cite{fioranelli2019radar}. Specifically, the attack model was fitted using 25 training samples and a subset of 25 unseen testing samples. The trained attack model was then used to infer data features to test their resilience to the black-box MIA. The prediction of the attack model based on the inferred features determine the membership status of a data point. 
For the rule-based black-box attack, the attacker did not fit the attack model but instead used predetermined rules to check the membership of a data point. To ensure a fair comparison, the same size of data used in the black-box MIA was used in the  rule-based black-box attack.

In creating the label-only ~\cite{choquette2021label}, we employed the HopSkipJump ~\cite{chen2020hopskipjumpattack} adversarial technique, a decision-based attack. The maximum query limit was set to two for calibrating the distance threshold during the generation of adversarial attacks. We began by creating an ART baseline model using the same architecture as the multi-task HAR model. This model was trained and tested with 20 samples over 3 epochs using a batch size of 128. The loss function and optimiser parameters were the same to those used in the baseline multi-task model. The ART model used 20 subset samples each for the training and testing to generate label-only attacks. Furthermore, a subset of 10 unseen samples were used to investigate the success rate of the generated attacks on the target HAR models (see Figure \ref{fig:labelonlymia}(a)).
Furthermore, to investigate the success rate of label-only attacks ~\cite{choquette2021label}, a larger number of samples were considered for the training and testing sizes in Figure \ref{fig:labelonlymia}(b). The attack training size was increased to 40, the testing size to 30, and the evaluation size to 20. The baseline ART model used in generating attacks in this scenario was trained at 10 epochs.

Furthermore three shadow models were initialized based on the multi-task HAR target model architecture. The dataset used to train and test the shadow models consisted of a subset of 25 data samples, attack training and testing size 25 each (out of 90 total samples available). Samples were drawn randomly out of the 25 samples for testing MIA black-box attack~\cite{shokri2017membership}. The learning rate and optimization parameters for each shadow model were kept consistent with those of the baseline multi-task HAR model, which were then used to create samples for testing the MIA black-box attack ~\cite{shokri2017membership}. To assess the success rate, a black-box attack was conducted to ascertain the membership status of the generated shadow data samples. 

\section{Results}
\subsection{HAR Performance of Attribution Based DP Models}
Table \ref{table:harcomp} presents the performance comparison of HAR model across various DP-based attribution methods tested on the Radar dataset ~\cite{fioranelli2019radar}. The comparison includes a baseline multi-task model without DP, a baseline learnable optics and other attribution methods with DP integration mentioned above, all using an equal value of $\epsilon = 1.20$ for consistency. Notably, adding noise while considering relevant attributions does not reduce model accuracy. In fact, performance accuracy improves across all attribution methods compared to the baseline model. IDG-DP achieves the best testing performance, followed by IG-DP, ISG-DP, and IIG-DP, respectively. The learnable optics method provides the lowest accuracy, likely due to the masking effect during model development. Since robustness is our primary concern, testing accuracy alone is insufficient to determine a method's effectiveness. It is also crucial to consider how the model behaves under MIA.

\begin{table}[h!]
\centering
\caption{Performance evaluation  comparison of HAR for the baseline and various multi-task DP based activity models. $\epsilon = 1.20$ attribution threshold = 0.00025 For Base-DP, IG-DP, Sal-DP, IIG-DP, ISG-DP and IDG-DP. noise mask for Optics = 0.50 }
\vspace{-1.5mm}
\small
\label{table:harcomp}
\begin{adjustbox}{width=0.37\textwidth}
\small
\begin{tabular}{|c|c|c|c|c|c|c|c|c|}
\hline
\multirow{2}{*}{Model}  & {Accuracy$\uparrow$}&{Precision$\uparrow$} & {Recall$\uparrow$} & {F1-Score$\uparrow$} \\
& {(\%)}  & {(\%)} & {(\%)}  & {(\%)}             \\ \hline
{Baseline}  &  {83.30} & {83.30} & {83.30} & {82.90}\\ \hline
{{Base-DP}}  &  {90.00} & {90.00} & {90.00} & {90.00} \\ \hline
{Optics} & {63.30} & {63.30} & {63.30}  & {56.10} \\ \hline
{IG-DP} & {93.30} & {93.30} & {93.30}  & {93.30}  \\ \hline
{GradC-DP}  & {83.30} & { 83.30} & {83.30} & {82.20} \\ \hline
{Sal-DP} & {80.00} & {80.00} & {80.00}  & {80.00} \\ \hline
{IIG-DP} & {90.00} & {90.00} & {90.00}  & {89.80} \\ \hline
{ISG-DP} & {90.00} & {90.00} & {90.00}  & {90.30} \\ \hline
\textbf{IDG-DP} & \textbf{96.70} & \textbf{96.70} & \textbf{96.70}  & \textbf{96.70} \\ \hline
\end{tabular}
\end{adjustbox}
\end{table}

\vspace{-4.5mm}
\subsubsection{Models Performance Against Black-box and Black-box Rule Based Attacks}\label{dp_results}
Table \ref{table:miabcomp} describes the performance of various models against black-box MIA ~\cite{shokri2017membership} on the Radar dataset ~\cite{fioranelli2019radar}. The metrics of total attack accuracy, precision, and recall were based on the chosen attack training and testing ratios. The test accuracy serves as the baseline clean accuracy target classifier, which is crucial for assessing model performance in the target estimator role.

The IDG-DP clean model accuracy of HAR is better as tested using the selected samples. The total attack recall for IGD-DP is lower than the baseline, indicating enhanced robustness, the GradC-DP, IIG-DP, and Sal-DP models show lower total attack recall. However, these models also exhibit reduced accuracy with the selected samples. Considering the IDG-DP clean model's performance with the HAR target model, it appears to be the preferable option as it effectively balances accuracy and attack resilience. This preference is further reinforced when evaluating against more robust adversarial black-box attacks designed to induce significant misclassification in the model. In summary, the IDG-DP model stands out as a promising approach, offering improved accuracy while maintaining strong defense against various attack strategies.

Table \ref{table:miarbcomp}, presents a comparative analysis of IDG-DP against  the baseline, optics, and other attribution-based DP methods againts blackbox rule-based MIA. IDG-DP achieves the highest clean  test accuracy among all tested methods. All tested methods outperform the baseline in mitigating privacy attacks, as evidenced by lower total attack accuracy and precision. The learnable optics method demonstrates the best performance in terms of recall, indicating strong resistance to privacy attacks. However, it also shows the lowest accuracy among all the techniques tested, highlighting a potential trade-off between privacy and utility.

\begin{table}[h!]
\caption{Performance evaluation comparison for the baseline and various multi-task DP based activity models against a black-box MIA. Attack training size = 25, attack test size = 25,  $\epsilon = 1.20$ attribution threshold = 0.00025 For Base-DP, IG-DP, Sal-DP, IIG-DP, ISG-DP and IDG-DP. noise mask for Optics = 0.50}
\vspace{-6.5mm}
\begin{center}
\label{table:miabcomp}
\begin{adjustbox}{width=0.42\textwidth}
\small
\begin{tabular}{|c|c|c|c|c|c|c|c|c|}
\hline
\multirow{2}{*}{Model}  & {Clean Test} & {Total Attack} & {Total Attack} & {Total Attack}   \\
& {Accuracy (\%)$\uparrow$}  & {Accuracy (\%)$\downarrow$} & {Precision (\%)$\downarrow$}  & {Recall (\%)$\downarrow$}             \\ \hline
{Baseline}  & {83.33} & {57.50} & {100.0} & {51.43} \\ \hline
{Base-DP}  & {90.00} & {57.50} & {100.0} & {51.43} \\ \hline
{Optics} & {63.33} & {66.70} & \textbf{66.70}  & {100.00} \\ \hline
{IG-DP} & {93.33} & {50.00} & {94.12}  & {45.71}  \\ \hline
{GradC-DP} & {83.33} & \textbf{45.00} & {88.24} & \textbf{42.86}  \\ \hline
{Sal-DP} & {80.00} & \textbf{45.00} & {88.24}  & \textbf{42.86}  \\ \hline
{IIG-DP} & {90.00} & {47.50} & {93.75}  & \textbf{42.86}  \\ \hline
{ISG-DP} & {90.00} & {52.50} & {94.44}  & {48.57}  \\ \hline
{IDG-DP} & \textbf{96.70} & {52.50} & {94.44}  & {48.57}  \\ \hline
\end{tabular}
\end{adjustbox}
\end{center}
\end{table}

\begin{figure*}
\centering
\subfloat[]{\includegraphics[scale = 0.1]{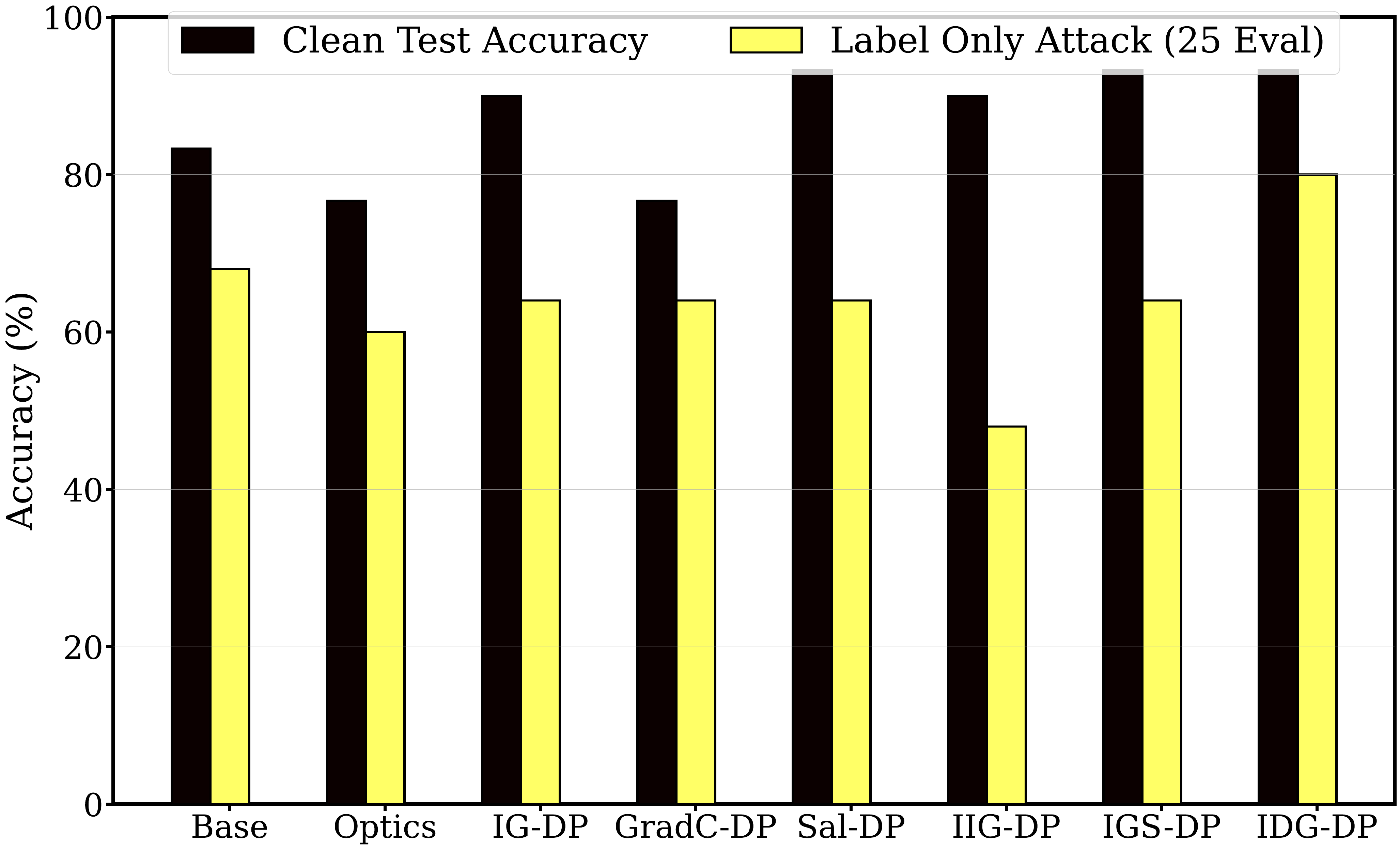}}
\subfloat[]{\includegraphics[scale = 0.1]{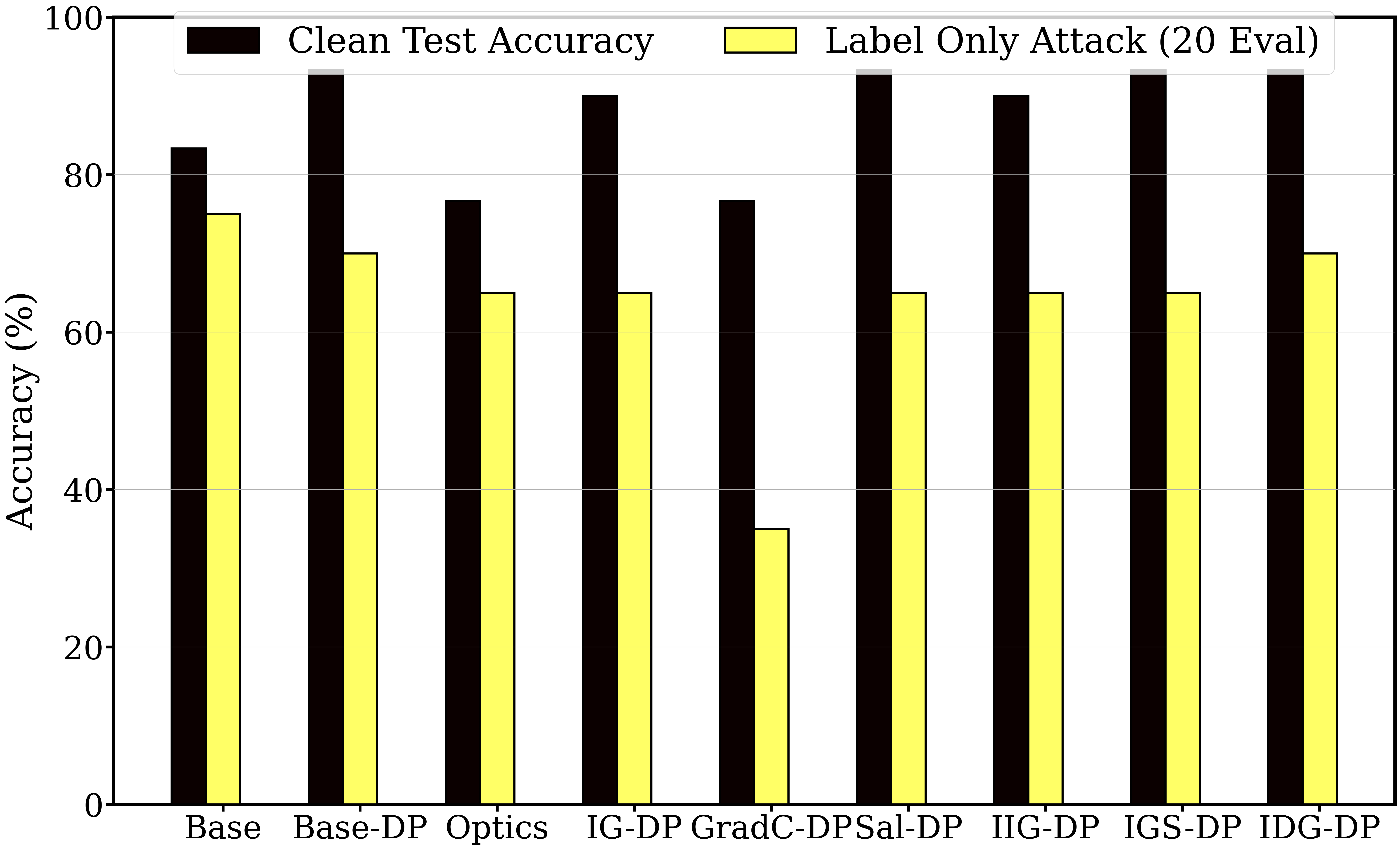}}
\vspace{-5.5mm}
\caption{Models performance comparison against Label-Only MIA: \textbf{a)} Label Only MIA, {attack training size = 25, attack test size = 25, evaluation size = 25},  $epsilon = 1.20$ attribution threshold = 0.00025 For {Base-DP} IG-DP, Sal-DP, IIG-DP, ISG-DP and IDG-DP. noise mask for Optics = 0.50, \textbf{b)} Label Only MIA, attack training size = 40, attack test size = 30, evaluation size = 20, $epsilon = 1.20$ attribution threshold = 0.00025 For IG-DP, Sal-DP, IIG-DP, ISG-DP and IDG-DP. noise mask for Optics = 0.50}
\label{fig:labelonlymia}
\end{figure*}


\vspace{-6.5mm}
\begin{table}[h!]
\caption{Performance evaluation comparison for the baseline and various multi-task DP based activity models against black-box rule based MIA. Attack training size = 25, attack test size = 25,  $\epsilon = 1.20$ attribution threshold = 0.00025 For Base-DP IG-DP, Sal-DP, IIG-DP, ISG-DP and IDG-DP. noise mask for Optics = 0.50}
\vspace{-6.5mm}
\begin{center}
\label{table:miarbcomp}
\begin{adjustbox}{width=0.42\textwidth}
\small
\begin{tabular}{|c|c|c|c|c|c|c|c|c|c|}
\hline
\multirow{2}{*}{Model}  & {Clean Test} & {Total Attack} & {Total Attack} & {Total Attack}   \\
& {Accuracy (\%)$\uparrow$} & {Accuracy (\%)$\downarrow$} & {Precision (\%)$\downarrow$}  & {Recall (\%)$\downarrow$}            \\ \hline
{Baseline}  & {83.33} &  {72.22} & {70.59} & {100.00}  \\ \hline
{Base-DP}  & {90.00} &  {70.00} & {68.97} & {100.00}  \\ \hline
{Optics} &  {63.33} & \textbf{60.00} & {69.35}  & \textbf{71.67}  \\ \hline
{IG-DP} & {93.30} & {68.89} & \textbf{68.18}  & {100.00}  \\ \hline
{GradC-DP}  & {80.00} & {66.67} & {69.23} & {90.00}  \\ \hline
{Sal-DP} & {80.00} & {66.67} & {69.23}  & {90.00}  \\ \hline
{IIG-DP} & {90.00} & {70.00} & {68.97}  & {100.00}  \\ \hline
{ISG-DP} & {90.00} & {68.69} & {68.60}  & {98.33}  \\ \hline
{IDG-DP} & \textbf{96.70} & {67.78} & {67.42}  & {100.00}  \\ \hline
\end{tabular}
\end{adjustbox}
\end{center}
\end{table}

\vspace{-5.5mm}
Table \ref{table:mibtsm}, presents the comparison results of IDG-DP against the baseline, optics, and other attribution-based DP methods, tested against black-box MIA \cite{shokri2017membership} using ten shadow models.  In each case, the clean testing accuracy of each tested technique were reported. Notably, IDG-DP exhibit superior performance across all metrics. This indicates enhanced protection against data leakage in the presence of black-box MIA, especially when the attacker lacks access to the original data samples.

\begin{table}[h!]
\caption{{Performance evaluation comparison for the baseline and various multi-task DP based activity models against black box MIA using 10 shadow models. Attack train size = 25, Attack test size = 25, Shadow dataset size = 25,  $epsilon = 1.20$ attribution threshold = 0.00025 For Base-DP, IG-DP, Sal-DP, IIG-DP, ISG-DP and IDG-DP. noise mask for Optics = 0.50}}
\vspace{-1.5mm}
\small
\centering
\label{table:mibtsm}
\begin{adjustbox}{width=0.47\textwidth}
\small
\begin{tabular}{|c|c|c|c|c|c|c|c|c|c|}
\hline
\multirow{2}{*}{Model}  & {Clean Test} & {Total Attack} & {Total Attack} & {Total Attack}   \\
& {Accuracy (\%)$\uparrow$} & {Accuracy (\%)$\downarrow$} & {Precision (\%)$\downarrow$}  & {Recall (\%)$\downarrow$}            \\ \hline
{Baseline}  & {83.33} &  {64.00} & {62.07} & {72.00}  \\ \hline
{Base-DP}  & {90.00} &  {56.00} & {55.17} & {64.00}  \\ \hline
{Optics} &  {63.33} & {66.00} & {75.00}  & {48.00}  \\ \hline
{IG-DP} & {93.33} & {44.00} & {41.18}  & {28.00}  \\ \hline
{GradC-DP}  & {80.00} & {58.00} & {57.14} & {64.00}  \\ \hline
{Sal-DP} & {80.00} & {42.00} & {44.44}  & {64.00}  \\ \hline
{IIG-DP} & {90.00} & {46.00} & {43.75}  & {28.00}  \\ \hline
{ISG-DP} & {90.00} & {48.00} & {46.67}  & {28.00}  \\ \hline
\textbf{IDG-DP} & \textbf{96.70} & \textbf{40.00} & \textbf{35.29}  & \textbf{24.00} \\ \hline
\end{tabular}
\end{adjustbox}
\end{table}




\vspace{-4.5mm}
\subsection{Models Performance Against Label-Only Adversarial Attacks}
To evaluate the robustness of the proposed IDG-DP method, we subjected it to more sophisticated, adversarial MIA. Specifically, the Label-Only ~\cite{choquette2021label} attack was employed, using the Radar ~\cite{fioranelli2019radar} dataset and the Hop-Skip-Jump ~\cite{chen2020hopskipjumpattack} adversarial perturbation technique to generate robust adversarial samples. The objective was to evaluate each method's resistance to this attack by observing their accuracy in detecting adversarial samples.

Figure \ref{fig:labelonlymia}(a) illustrates the results of a Label-Only ~\cite{choquette2021label} attack with the following parameters: i) attack training size equal to 25 samples, ii) attack testing size equal to 25 samples and iii) evaluation samples equal to 25 randomly selected. IDG-DP consistently outperforms other methods in both clean test accuracy and attack test accuracy. Notably, the label-only attack significantly impacted the baseline and Base-DPmodels, reducing their accuracy from 83.33\%, 93.33\%  to 68.0\%, respectively. 

Figure \ref{fig:labelonlymia}(b) illustrates the performance against a more extensive Label-Only ~\cite{choquette2021label} attack generated with the following parameters: i) attack training size equal to 40 samples, ii) attack testing size equal to 30 samples and iii) evaluation samples equal to 20 samples randomly selected.  In all cases, IDG-DP outperforms all tested attribution methods in resisting the attack. While the baseline model slightly outperformed IDG-DP in attack resistance, it achieved this at the expense of lower clean test accuracy, compromising the model's utility.

\section{Conclusion}
Our study breaks new ground in addressing privacy preservation for radar-based human motion analysis models. 
We introduce a novel IDG-DP method, which exploits the solid theoretical guarantees of Differential Privacy (DP) with a principle way driven by the Integrated Decision Gradients (IDG) of a multi-task model to preserve the utility of the dataset. We assess IDG-DP's robustness against various black-box membership inference attacks (MIA), including sophisticated label-only attacks. Our findings demonstrate that DP, when applied with carefully selected attributions effectively mitigates MIA risks. IDG-DP stands out for its ability to maintain strong HAR performance, while effectively countering MIA attacks. In the attack target HAR model, IDG-DP's accuracy surpasses that of the tested benchmarks, making it an optimal choice for balancing privacy protection and utility in activity recognition. 
We plan to test our methods on more extensive and diverse datasets to confirm their scalability, resistance and generalizability. 
This research provides a solid foundation for developing privacy-preserving techniques in radar-based human motion analysis, paving the way for more secure and ethical applications in healthcare monitoring and activity recognition.

\vspace{-2.5mm}
\section{Acknowledgments}

We acknowledge funding from EPSRC(EP\verb|\|W01212X \verb|\|1), the Royal Society (RGS\verb|\|R2\verb|\|212199) and Academy of Medical Sciences (NGR1\verb|\|1678).


{\small
\bibliographystyle{ieee_fullname}
\bibliography{hc_wacv_camera_ready.bib}
}

\clearpage
\onecolumn
\appendix

\section{Multi-Task Network Architecture}\label{multi-task-arch}

A multi-task network that is able to perform both HAR and subject identification has been constructed. Its architecture is summarised in Table \ref{architecture}. The backbone ResNet18 convolutional layers was used to extract features from each sample. Subsequently, these features are fed into two branches, allowing for a sample's subject and activity classification. 
\renewcommand\thetable{\thesection.\arabic{table}} 
\setcounter{table}{0}
\vspace{-3.5mm}
\begin{table}[ht]
\small
\centering
\begin{adjustbox}{width=0.42\textwidth}
\small
\begin{tabular}{|p{1.8cm}|p{1.7cm}|p{3.65cm}|}
\hline
\emph{Layer Name} & \emph{Output Size} & \emph{Description} \\
\hline
\hline
 \multicolumn{3}{|l|}{\textbf{ResNet Features Extraction}}\\
\hline
Conv2d & 64 & 7x7 stride 2\\
\hline
MaxPool2d & 64 & 3x3 stride 1\\
\hline
4 x Conv2d & 64 & 3x3 stride 1\\
\hline
4 x Conv2d & 128 & 3x3 stride 2\\ 
\hline
4 x Conv2d & 256 & 3x3 stride 1\\ 
\hline
\multicolumn{3}{|l|}{\textbf{1. Subject Branch}}\\
\hline
4 x Conv2d & 512 & 3x3 stride 1\\\hline
AvgPool2d  & 512 & Adaptive average pooling \\\hline
Flatten  & 512 &  Convert to a vector\\\hline
Linear & 5/10 &  Fully connected layer\\\hline
 \multicolumn{3}{|l|}{\textbf{2. Activity Branch}}\\\hline
4 x Conv2d & 512 & 3x3 stride 1\\\hline
AvgPool2d  & 512 & Adaptive average pooling \\\hline
Flatten  & 512 &  Convert to a vector\\\hline
Linear & 3 &  Fully connected layer\\\hline
\end{tabular}
\end{adjustbox}
\caption{\label{architecture}\textbf{Multi-Task Network Architecture Summary.} }
\end{table}
\vspace{-3mm}

\section{Effects of Attribution Threshold and Epsilon for Sal, GradC, IG, IIG and ISG}\label{mt_performance}

Figure ~\ref{figure:salatr}, Figure ~\ref{figure:gradatr}, Figure  ~\ref{figure:igatr}, Figure  ~\ref{figure:iigatr}, Figure ~\ref{figure:isgatr} show the impact of various $\epsilon$ and attribution threshold values on the performance of user and activity recognition using the multi-task model. DP noise is driven by the attribution methods Sal, GradC, IG, IIG, ISG, respectively.
\renewcommand\thefigure{\thesection.\arabic{figure}} 
\setcounter{figure}{0}  
\begin{figure}[H]
\centering
\vspace{-3.5mm}
\subfloat[\:$\epsilon$ and Attribution Thresholds performance]{\includegraphics[width=0.30\textwidth]{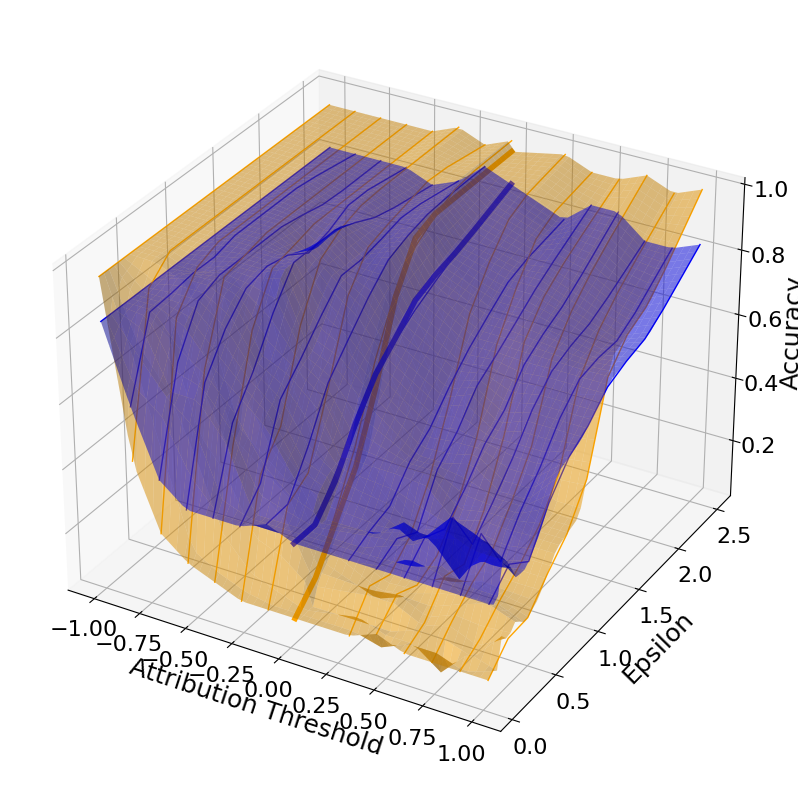}}
\subfloat[\:Attribution Threshold $0.00025$ against $\epsilon$]{\includegraphics[width=0.35\textwidth]{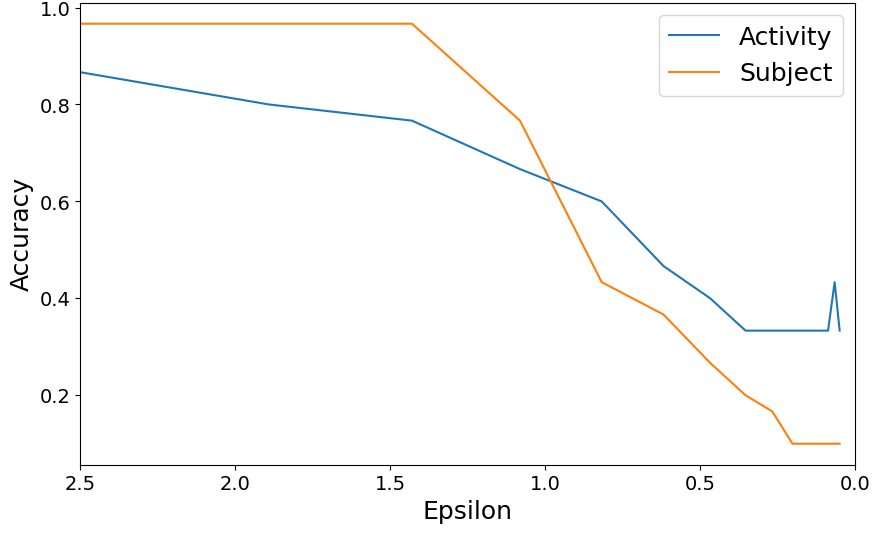}}
\subfloat[\:Pixel Attribution for Subject and Activity]{{\includegraphics[width=0.32\textwidth]{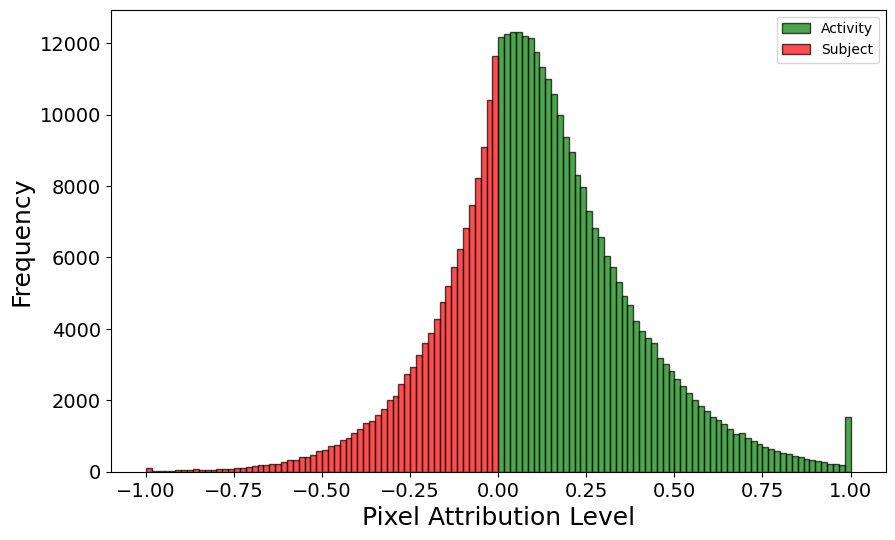}}}
\caption{\textbf{Impact of Different Saliency (Sal) Attribution Thresholds and  $\epsilon$ Levels on the performance of HAR and subject recognition.} a) Multi-task models performance across various $epsilon$ and attribution threshold values, b) Multi-task models performance across various $epsilon$ with attribution threshold equal to 0.00025, c) Histogram of pixel attributions for Subject and Activity recognition.}
\label{figure:salatr}
\end{figure}
\vspace{-3.0mm}
\setcounter{figure}{1} 
\begin{figure}
\centering
\subfloat[\:$\epsilon$ and Attribution Thresholds performance]{\includegraphics[width=0.30\textwidth]{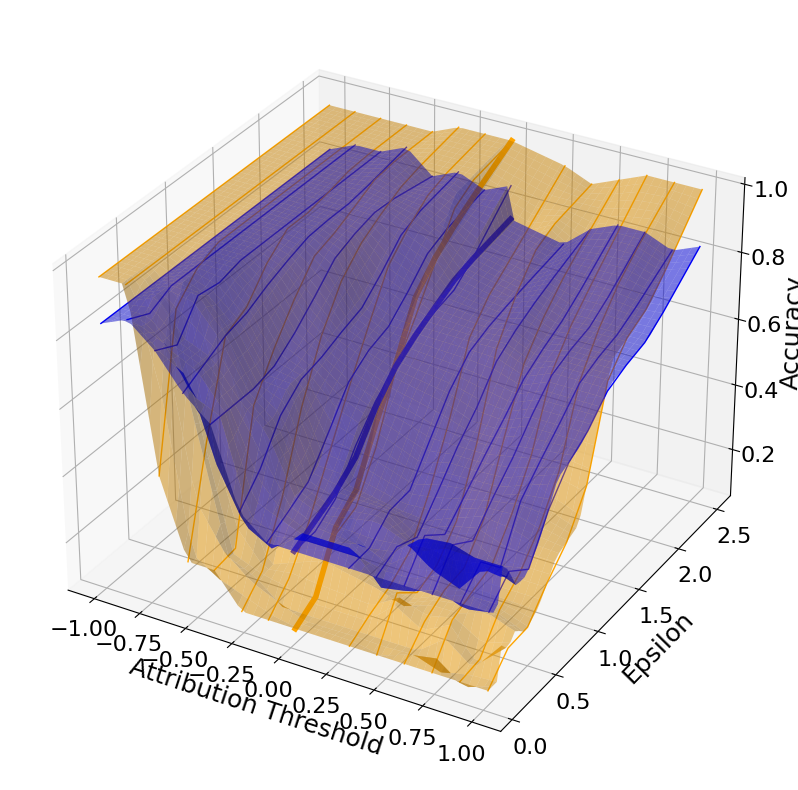}}
\subfloat[\:Attribution Threshold $0.00025$ against $\epsilon$]{\includegraphics[width=0.35\textwidth]{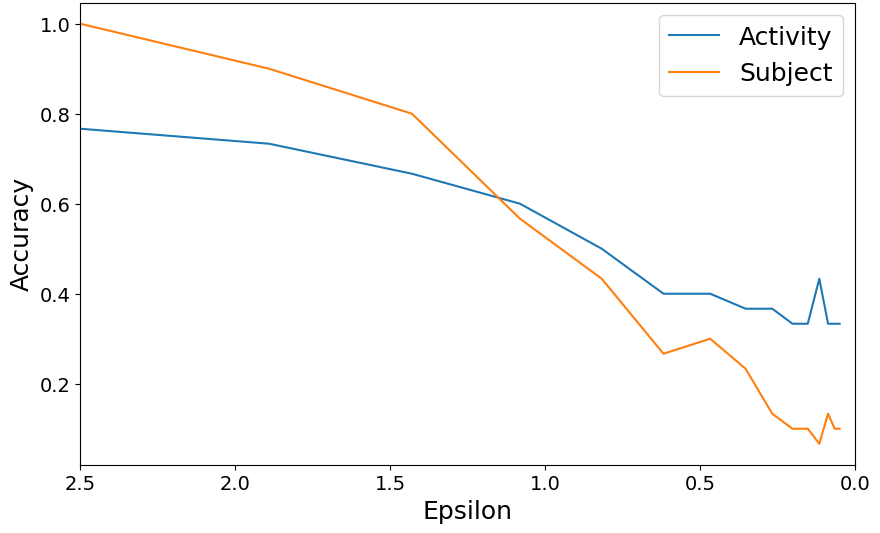}}
\subfloat[\:Pixel Attribution for Subject and Activity]{{\includegraphics[width=0.32\textwidth]{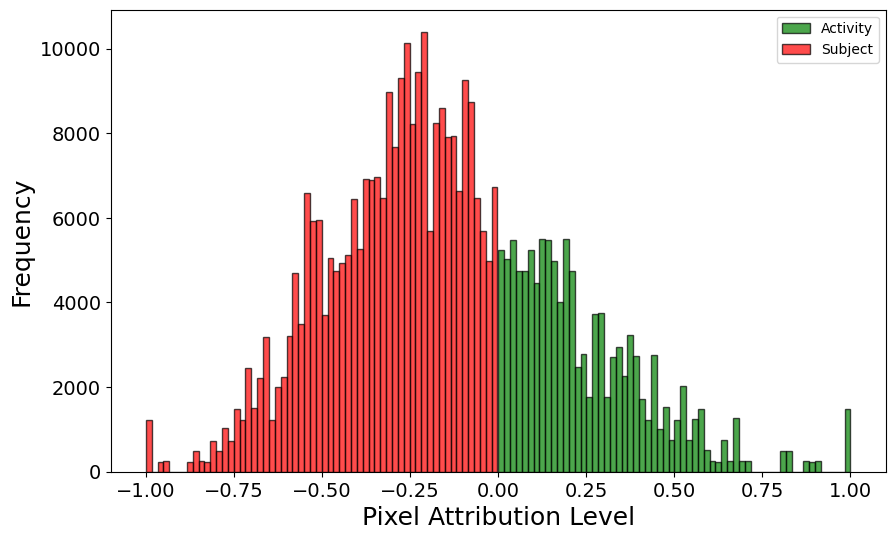}}}
\caption{\textbf{Impact of Different Gradcam (GradC) Attribution Thresholds and  $\epsilon$ Levels on the performance of HAR and subject recognition.} a) Multi-task models performance across various $epsilon$ and attribution threshold values, b) Multi-task models performance across various $epsilon$ with attribution threshold equal to 0.00025, c) Histogram of pixel attributions for Subject and Activity recognition.}
\label{figure:gradatr}
\end{figure}
\vspace{-3.5mm}

\setcounter{figure}{2} 
\begin{figure}
\centering
\subfloat[\:$\epsilon$ and Attribution Thresholds performance]{\includegraphics[width=0.30\textwidth]{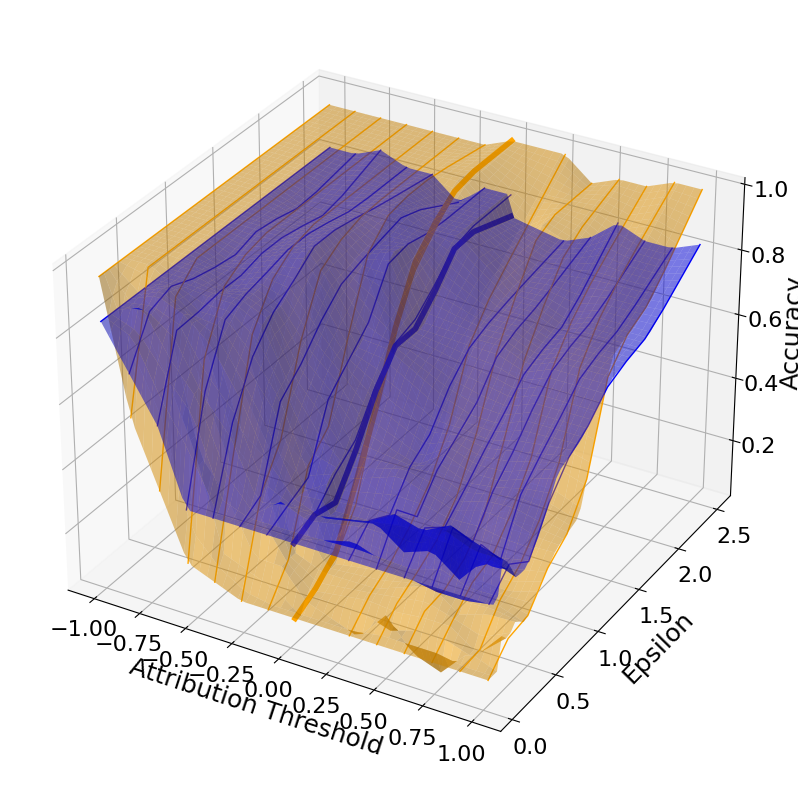}}
\subfloat[\:Attribution Threshold $0.00025$ against $\epsilon$]{\includegraphics[width=0.35\textwidth]{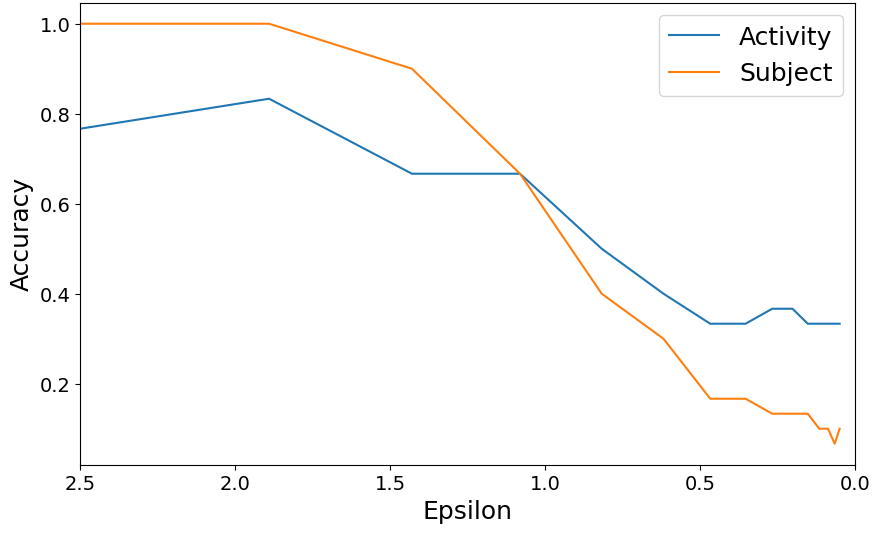}}
\subfloat[\:Pixel Attribution for Subject and Activity]{{\includegraphics[width=0.32\textwidth]{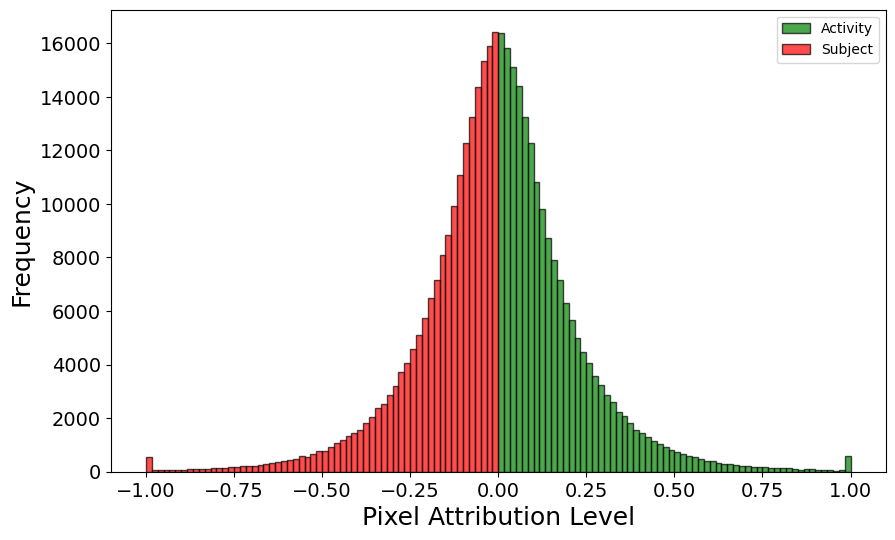}}}
\caption{\textbf{Impact of Different Integrated Gradient (IG) Attribution Thresholds and  $\epsilon$ Levels on the performance of HAR and subject recognition.} a) Multi-task models performance across various $epsilon$ and attribution threshold values, b) Multi-task models performance across various $epsilon$ with attribution threshold equal to 0.00025, c) Histogram of pixel attributions for Subject and Activity recognition.}
\label{figure:igatr}
\end{figure}
\vspace{-3.5mm}

\setcounter{figure}{3} 
\begin{figure}[H]
\centering
\subfloat[\:$\epsilon$ and Attribution Thresholds performance]{\includegraphics[width=0.30\textwidth]{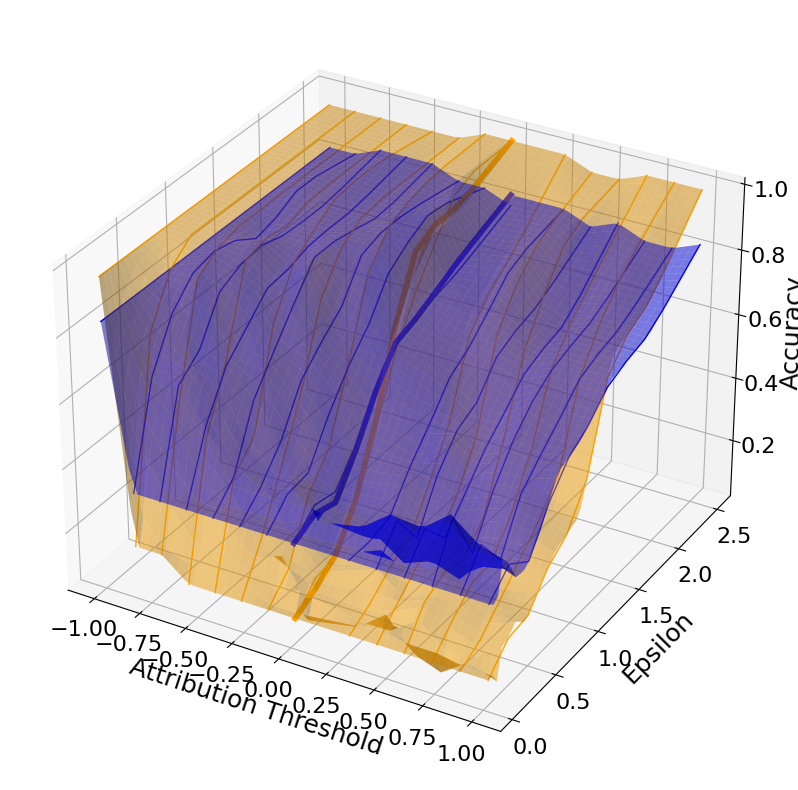}}
\subfloat[\:Attribution Threshold $0.00025$ against $\epsilon$]{\includegraphics[width=0.35\textwidth]{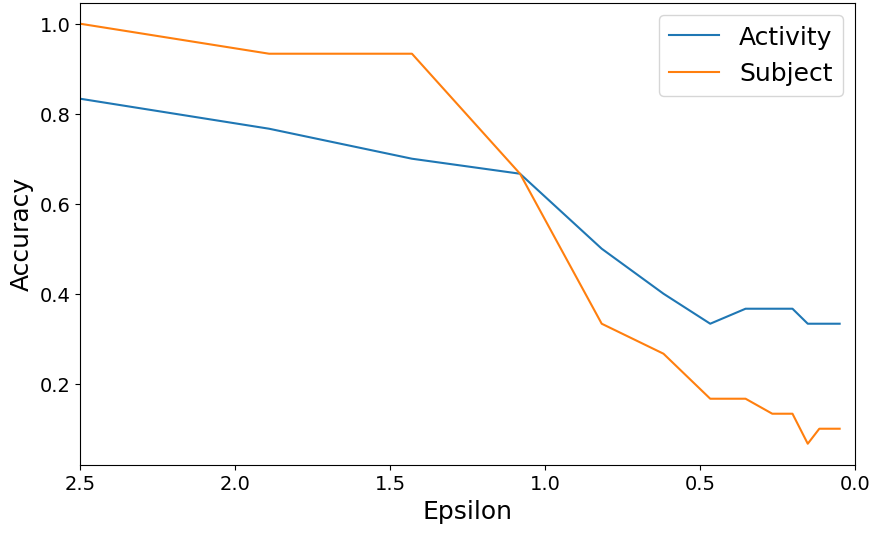}}
\subfloat[\:Pixel Attribution for Subject and Activity]{{\includegraphics[width=0.32\textwidth]{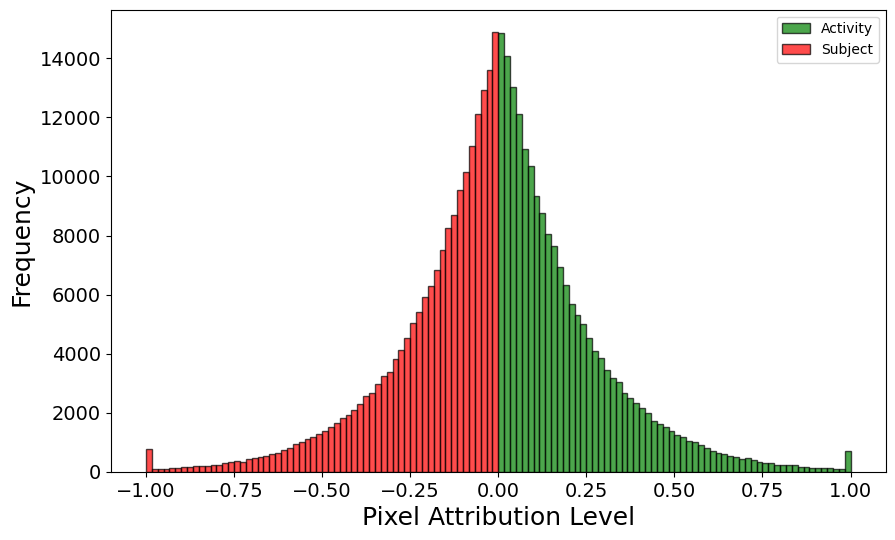}}}
\caption{\textbf{Impact of Different Integrated InputX Gradient (IIG) Attribution Thresholds and  $\epsilon$ Levels on the performance of HAR and subject recognition.} a) Multi-task models performance across various $epsilon$ and attribution threshold values, b) Multi-task models performance across various $epsilon$ with attribution threshold equal to 0.00025, c) Histogram of pixel attributions for Subject and Activity recognition }
\label{figure:iigatr}
\end{figure}
\vspace{-3.5mm}

\setcounter{figure}{4} 
\begin{figure}[H]
\centering
\subfloat[\:$\epsilon$ and Attribution Thresholds performance]{\includegraphics[width=0.30\textwidth]{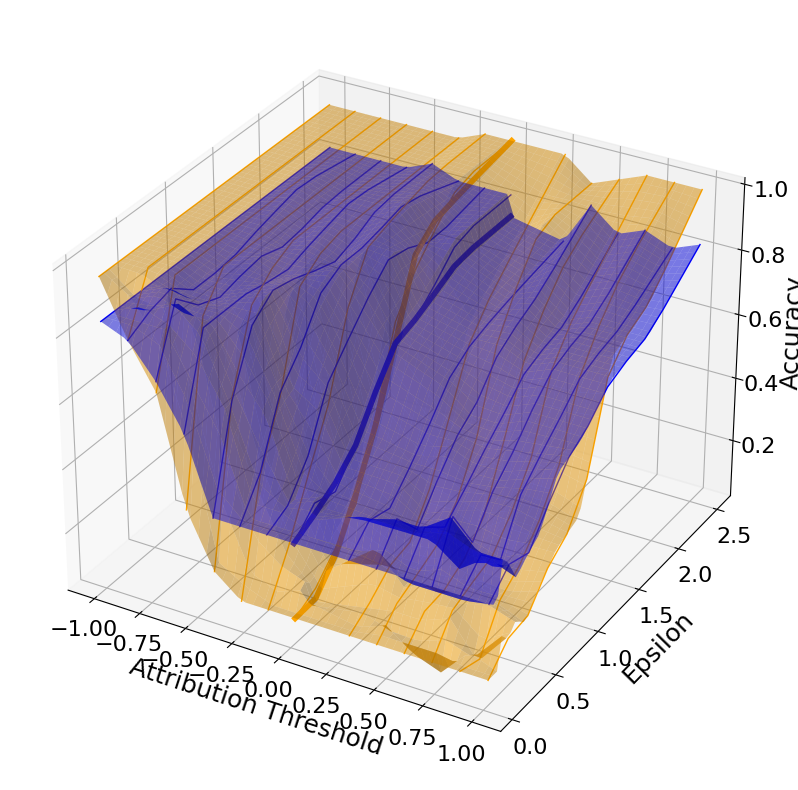}}
\subfloat[\:Attribution Threshold $0.00025$ against $\epsilon$]{\includegraphics[width=0.35\textwidth]{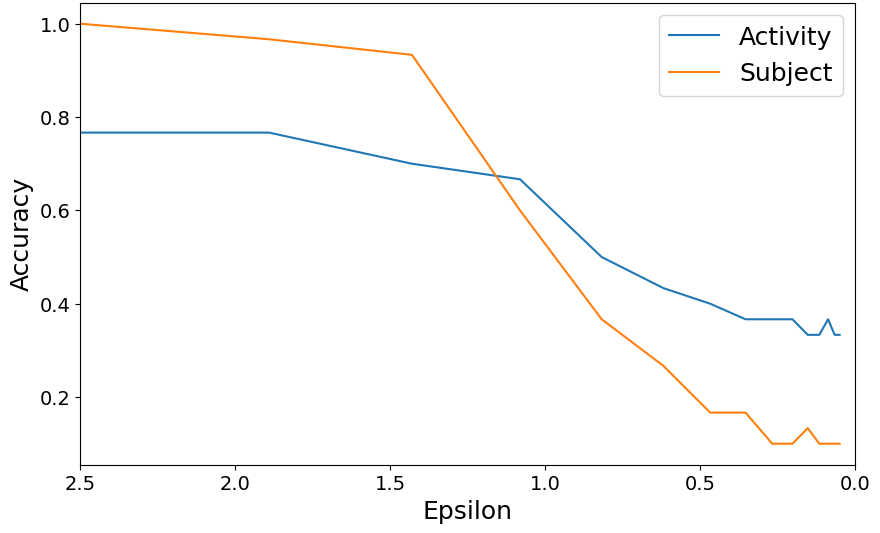}}
\subfloat[\:Pixel Attribution for Subject and Activity]{{\includegraphics[width=0.32\textwidth]{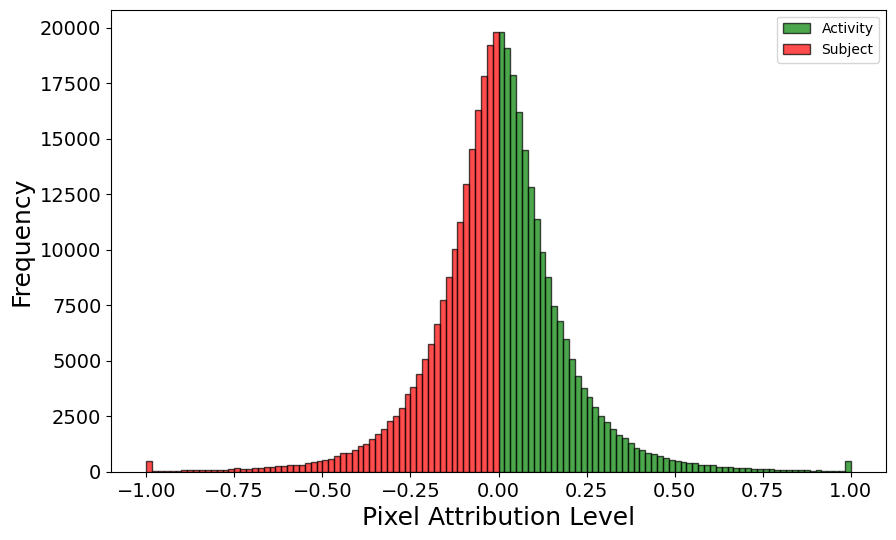}}}
\caption{\textbf{Impact of Different Integrated Integrated Smooth Gradient (ISG) Attribution Thresholds and  $\epsilon$ Levels on the performance of HAR and subject recognition.} a) Multi-task models performance across various $epsilon$ and attribution threshold values, b) Multi-task models performance across various $epsilon$ with attribution threshold equal to 0.00025, c) Histogram of pixel attributions for Subject and Activity recognition.}
\label{figure:isgatr}
\end{figure}
\vspace{-3.5mm}

\section{Implementation Details} \label{implem_details}

For a fair comparison, the multi-task/HAR baseline model, Optics, IDG-DP and other tested DP-based attribution methods, were trained with 32 batch size, 400 epochs using cross-entropy loss and a Stochastic Gradient Descent (SGD) optimizer with a learning rate of 0.001 and a momentum value of 0.9. An early stopping mechanism with a patience of 8 epochs and a minimum delta of 0.01 was implemented to ensure better model convergence. The Captum library ~\cite{kokhlikyan2020captum} was used for generating attributions, except for the IDG where the saliency function in \footnote{\url{https://github.com/chasewalker26/Integrated-Decision-Gradients}} was utilized to generate attributions. 

For the optic masking, the convolutional 2D layer was used to generate masks, the noise strength value used is 0.5, this mask is added to the original input to create a noisy input. The optic mask model were trained using 600 epochs with the same optimizer, momentum, learning rate and loss function with the baseline model.

Table \ref{table:data} describes the number of samples used for each evaluation procedure. The 60 {blue}{and 25} samples used in the shadow model black-box MIA are for generating member and non-member data to evaluate the success rate of the MIA attack. Additionally, Label-only-10 has 10 samples for evaluation, while Label-only-20 consists of 20 samples for evaluating the success rate of the attack.

\begin{table}[h!]
    \caption{Distribution of training and testing samples for each implementation procedure.}
    \centering
    \label{table:data}
    \begin{tabular}{|c|c|c|}
        \hline
        {Procedure}& {Training} & {Testing}   \\ \hline
 {HAR}  & {60} & {30}   \\ \hline
 {Blackbox MIA}  & {25} & {25}   \\ \hline
 {Rule-based MIA}  & {25} & {25}   \\ \hline
 {{Blackbox MIA with 3 shadow model}}  & {60} & {30}   \\ \hline
 {{Blackbox MIA with 10 shadow model}}  & {{25}} & {{25}}  \\ \hline
 {{Label-Only 25}}  & {{25}} & {{25}} \\ \hline
 {Label-Only 10}  & {20} & {20}   \\ \hline
 {Label-Only 20}  & {40} & {30}   \\ \hline
 \multicolumn{3}{l}{*Evaluation samples for Label-Only 10 = 10, and 20 for Label-Only 20.}
    \end{tabular}   
\end{table}
\vspace{-3mm}

\section{Performance Comparison with the inclusion of Baseline DP (Base-DP) against HAR and tested attacks} \label{base_dp_performance}

Laplace noise was introduced to the baseline multi-task model to create the Base-DP model (without attributions guidance) for comparative analysis with the tested methods. An attribution threshold of $0.00025$ and an $\epsilon$ value of 1.20 were applied in constructing the Base-DP model. 

For completeness, we added the HAR performance of Base-DP along with the performance of various attacks on the Base-DP model on all the results presented in the Results section. 

The investigation results in Table ~\ref{table:mibdpsm} demonstrate IDG's effective utility and resistance to black-box MIA attacks based on shadow models. IDG {demonstrate better performance compared with}  tested methods, including the baseline with DP {especially in terms of utility, attack accuracy and precision}. 

\begin{table}[H]
\caption{Performance evaluation comparison for the baseline and various multi-task DP based activity models against black-box MIA using shadow models. Shadow dataset size = 60,  $epsilon = 1.20$ attribution threshold = 0.00025 For Base-DP, IG-DP, Sal-DP, IIG-DP, ISG-DP and IDG-DP. noise mask for Optics = 0.50}
\vspace{-1.5mm}
\small
\centering
\label{table:mibdpsm}
\begin{adjustbox}{width=0.49\textwidth}
\small
\begin{tabular}{|c|c|c|c|c|c|c|c|c|c|}
\hline
\multirow{2}{*}{Model}  & {Clean Test} & {Total Attack} & {Total Attack} & {Total Attack}   \\
& {Accuracy (\%)$\uparrow$} & {Accuracy (\%)$\downarrow$} & {Precision (\%)$\downarrow$}  & {Recall (\%)$\downarrow$}            \\ \hline
{Baseline}  & {83.33} &  {56.67} & {54.17} & {86.67}  \\ \hline
{Base-DP}  & {90.00} &  {56.67} & {55.00} & {73.33}  \\ \hline
{Optics} &  {63.33} & {63.33} & {64.29}  & {60.00}  \\ \hline
{IG-DP} & {93.33} & {56.67} & {55.00}  & {73.33}  \\ \hline
{GradC-DP}  & {80.00} & {56.67} & {54.17} & {86.67}  \\ \hline
{Sal-DP} & {80.00} & {56.67} & {55.00}  & {73.33}  \\ \hline
{IIG-DP} & {90.00} & \textbf{36.67} & \textbf{37.50}  & \textbf{40.00}  \\ \hline
{ISG-DP} & {90.00} & {46.47} & {45.45}  & \textbf{33.33}  \\ \hline
{IDG-DP} & \textbf{96.70} & \textbf{36.67} & {37.50}  & {40.00} \\ \hline
\end{tabular}
\end{adjustbox}
\end{table}

\begin{figure}[H]
\centering
\small
{\includegraphics[width=0.49\textwidth]{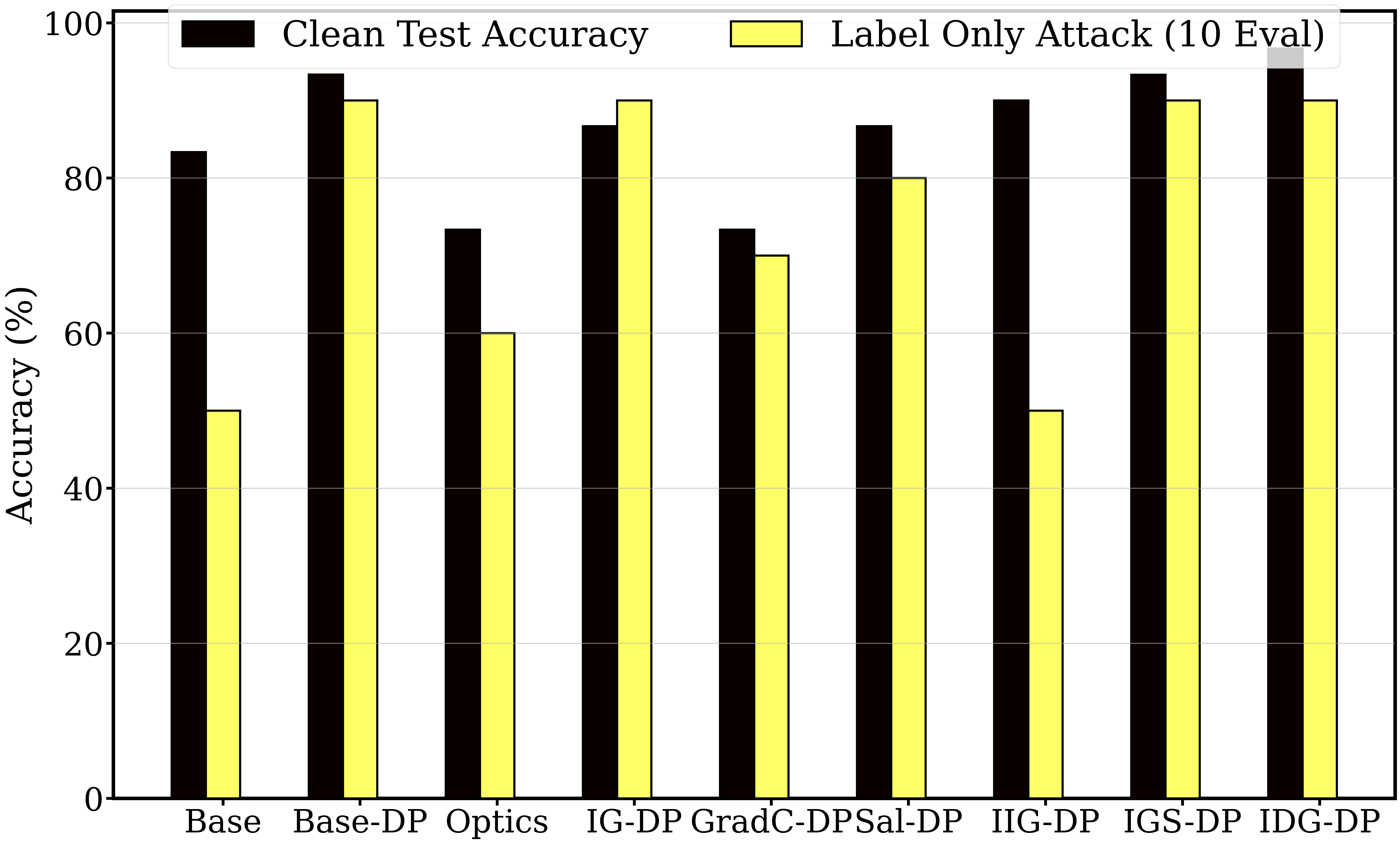}}
\vspace{-1.5mm}
\caption{Models performance comparison against Label-Only MIA:  {Label Only MIA, attack training size = 20, attack test size = 20, evaluation size = 10},  $epsilon = 1.20$ attribution threshold = 0.00025 For Base-DP, IG-DP, Sal-DP, IIG-DP, ISG-DP and IDG-DP. noise mask for Optics = 0.50} 
\label{fig:labelonlybdpmia}
\end{figure}

{Figure ~\ref{fig:labelonlybdpmia} (a) illustrates the results of a Label-Only attack~\cite{choquette2021label} evaluated using the data distribution for Label-Only 25} as reported in Table ~\ref{table:data}. IDG-DP consistently outperforms {all tested} methods in terms of clean test accuracy. In attack accuracy, IDG-DP and Base-DP exhibit similar performance, yet IDG-DP demonstrates superior utility. Figure \ref{fig:labelonlybdpmia} further demonstrates the utility and attack performance of IDG-DP and Base-DP for Label-Only 20, with additional data records provided in Table ~\ref{table:data}.

\vspace{-3mm}


\end{document}